\pdfoutput=1
\documentclass[acmsmall]{acmart}
\renewcommand\footnotetextcopyrightpermission[1]{}
\usepackage{pifont,adjustbox,makecell,longtable,arydshln,multicol,tabularx,algorithm,multirow,tcolorbox,subcaption,algpseudocode,colortbl}
\begin{document}
\title{RoSci: Exploring Robustness in Science VLMs}
\newcommand{\affnum}[1]{\textsuperscript{#1}}

\author{Junying Wang\affnum{1,2}, Yijin Guo\affnum{2,3},
  Yalun Wu\affnum{3}, Ziheng Jiao\affnum{2},
  Farong Wen\affnum{2,3}, Qihang Yan\affnum{2},
  Wenzhe Li\affnum{2}, Ye Shen\affnum{2,3},
  Qi Jia\affnum{2}, Chunyi Li\affnum{2,3,4},
 Zicheng Zhang\affnum{2,\textasteriskcentered}, Guangtao Zhai\affnum{2,3}}
\authornote{Corresponding authors: Zicheng Zhang
  (zhangzicheng@pjlab.org.cn) and Guangtao Zhai
  (zhaiguangtao@pjlab.org.cn).}

\affiliation{\institution{\affnum{1}Fudan University}\country{China}}
\affiliation{\institution{\affnum{2}Shanghai AI Laboratory}\country{China}}
\affiliation{\institution{\affnum{3}Shanghai Jiao Tong University}\country{China}}
\affiliation{\institution{\affnum{4}Nanyang Technological University}\country{Singapore}}

\renewcommand{\shortauthors}{Wang et al.}
\begin{abstract}
Recent advancements in Vision-Language Models (VLMs) enable significant progress in scientific tasks. However, these models are generally assumed to be presented with complete and accurate information, overlooking the challenges of erroneous/missing information in wild scenarios. To address this gap, we introduce the concept of \textbf{Scientific Model Robustness}, which defines a model's ability to detect anomalies, assess solvability, and perform self-calibration when faced with \textit{Information Error} (inconsistencies) or \textit{Information Loss} (insufficiency). Specifically, we present \textbf{RoSci-Bench}, a diagnostic benchmark of 1,220 samples using a {seed-variant} strategy across six degradation types. To address VLM fragility, we further propose \textbf{RoSci-Agent}, a framework that employs a {Parse-Verify-Decide} pipeline to vet premises before reasoning. The evaluated VLMs sometimes answer questions whose premises are invalid or incomplete. RoSci-Agent achieves a higher Overall Robustness Score than the evaluated standalone VLMs. RoSci-Bench gives the multimodal reasoning community a common test of when models answer, request clarification, or reject a scientific question.
\end{abstract}
\authorsaddresses{}
\maketitle
\begin{figure}
    \centering
    \vspace{-0.1cm}
    \includegraphics[width=0.85\linewidth]{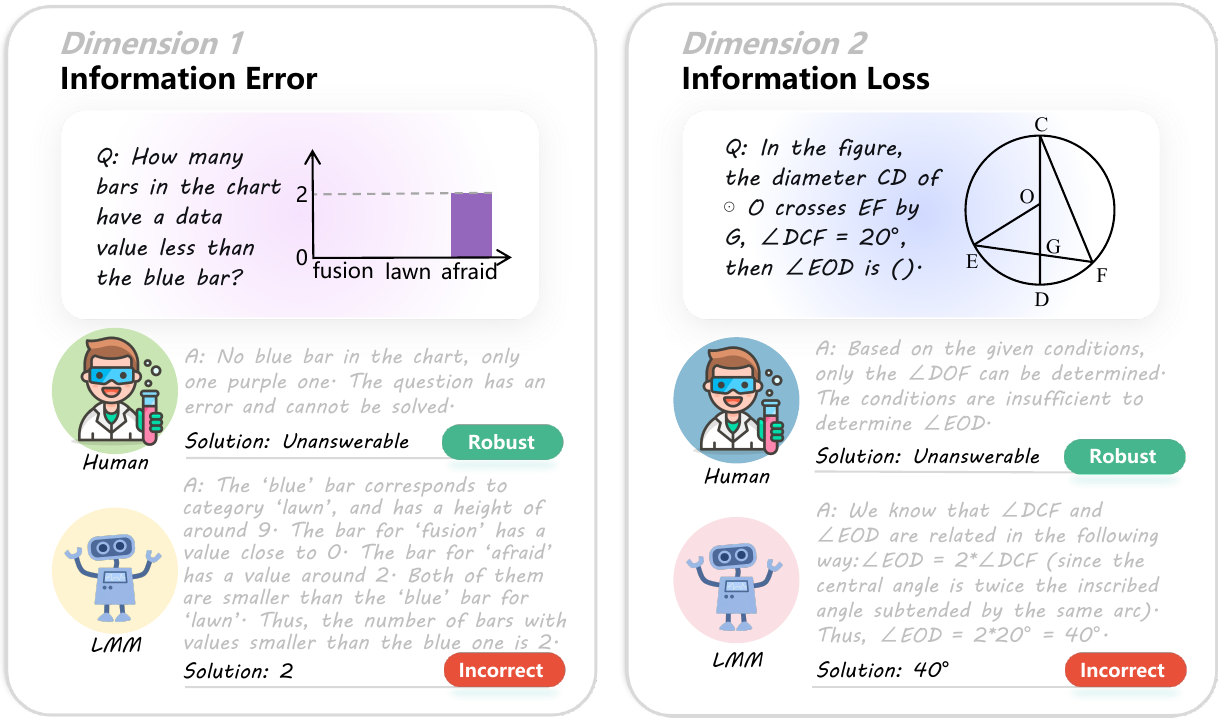}
    \vspace{-0.4cm}
    \Description{Two questions contrast human judgments with model answers. At left, a chart contains a single purple bar, but the question asks about a blue bar. The human flags the missing referent. The model invents a blue bar and answers two. At right, a circle diagram gives a 20-degree angle. The human explains that the requested angle cannot be determined. The model doubles the given angle and answers 40 degrees.}
    \caption{Illustration of Scientific Model Robustness dimensions: Information Error (left) and Information Loss (right). Note that even when a query is judged as unanswerable, the VLM still attempts to provide an incorrect response based on hallucinated premises.}
    \vspace{-0.4cm}
    \label{fig:moti}
\end{figure}

\section{Introduction}
\label{sec:intro}
Vision-Language Models (VLMs) continue to make significant progress in multi-modal scientific reasoning tasks \cite{AIBench,zhang2025large,zhang2025qeval100k}. Recent benchmarks evaluate VLMs on scientific reasoning and multimodal question answering \cite{mmmu,eese,qmirror,Adv-CPG}. These tasks include chart interpretation and cross-modal reasoning. However, existing evaluation tasks generally assume that models are presented with complete and accurate information, overlooking the \textit{common challenges of erroneous and missing information that frequently arise in real-world scientific scenarios}. As shown in Figure \ref{fig:moti}, in scientific reasoning, when faced with data inconsistencies or incomplete information, human researchers often pause their reasoning, recheck assumptions, or question the provided conditions. We examine whether VLMs perform similar checks when scientific inputs are incomplete or inconsistent.

To address this, we introduce a new research concept: \textbf{Scientific Model Robustness}. This concept aims to represent a model’s ability to identify anomalies, assess solvability, and perform self-calibration in scientific reasoning tasks. We evaluate robustness along two dimensions: 1) \textbf{Information Error Robustness}, which indicates the model’s ability to recognize factual errors, logical inconsistencies, or cross-modal contradictions and reject unreasonable inferences, and 2) \textbf{Information Loss Robustness}, which stands for the model’s ability to identify missing or incomplete conditions and express uncertainty or abstain instead of making unsupported conclusions.

To systematically evaluate \textbf{Scientific Model Robustness}, we develop a comprehensive evaluation framework. Specifically, we establish a systematic evaluation framework comprising 1,220 tasks. Starting with 305 expert-curated seeds, we employ a controlled augmentation pipeline to generate three semantically distinct variants for each seed. The variants probe how models respond to changes in wording or visual context while the underlying challenge is preserved. These variants simulate real-world challenges, such as erroneous data or missing information, which are common in scientific settings. The degradation is divided into two main categories: 1) Information Error Insertion, where factual inaccuracies, methodological violations, or cross-modal contradictions are introduced into the text or images without altering the core task, and 2) Information Loss Injection, where key information in the text or images is deleted, obscured, or weakened, which can leave the query without sufficient information for a unique answer. We evaluate reasoning stability, rejection F1, justification quality, and overall robustness.

To address the identified limitations, we propose a Robustness Agent System, which is designed to enhance the model's ability to handle uncertain or degraded inputs. This agent operates through a three-step process: 1) Solvability Detection, where the agent first analyzes the task to identify the goal, known conditions, and constraints, 2) Anomaly Recognition, where the agent detects inconsistencies, contradictions, or missing information between the textual and visual input, and 3) Selective Prediction, where the agent applies a decision rule based on the detected anomalies. If the agent determines that the input contains significant errors or missing information, it either rejects the task or requests clarification. Otherwise, it proceeds with reasoning and generates an answer. The agent checks the premises before generating an answer. Our contributions are as follows:

\begin{itemize}
\item The introduction of the concept of Scientific Model Robustness, which is defined through two core dimensions: Information Error Robustness and Information Loss Robustness.
\item The development of a scientific robustness benchmark, which employs systematic text and visual degradation strategies to construct degraded seeds and their variants from valid source problems. This benchmark facilitates reproducible evaluations of scientific model robustness, highlighting the vulnerabilities of current models when confronted with incomplete or erroneous inputs.
\item The proposal of the RoSci-Agent, an active mechanism designed to enable models to perform solvability detection, anomaly recognition, and selective prediction.
\end{itemize}

\section{Related Work}
We review advances in scientific reasoning with VLMs under idealized conditions (\S~\ref{subsec2.1}), limitations of current robustness evaluations (\S~\ref{subsec2.2}), and methods for improving VLM reliability that motivate our work (\S~\ref{subsec2.3}).

\subsection{Scientific Reasoning in Idealized Conditions}
\label{subsec2.1}
VLMs demonstrate remarkable capabilities in multi-modal scientific reasoning. This progress is largely driven by benchmarks like {ScienceQA} \cite{lu2022learn}, {MMMU} \cite{mmmu}, {MathVista} \cite{mathvista}, and {ChartQA} \cite{chartqa}, which advance model capabilities in complex chain-of-thought reasoning, expert-level knowledge understanding, and chart parsing. We evaluate models under a complementary condition in which a premise is corrupted or omitted.

\subsection{Robustness Evaluation Paradigm}
\label{subsec2.2}
Prior work on VLM robustness examines both perceptual and semantic changes to the input. One line of work focuses on {perceptual robustness}. For instance, the {RDR} framework \cite{rdr} investigates the impact of diagram noise and occlusions, while {TextVQA-C} \cite{textvqa-c} injects visual corruptions. Such research addresses `signal-level' distortions, which, while important, is tangential to the `logic-level' validity central to science. A more relevant line of inquiry addresses {semantic robustness}, with works like the {MMR} Benchmark \cite{mmr} tackling leading questions and {JBA} \cite{jba} constructing false-premise queries. RoSci-Bench organizes six degradation types under information error and information loss, including contradictions between text and images.

\subsection{Reliability Enhancement Strategies}
\label{subsec2.3}
To enhance model reliability against anomalous inputs, researchers propose various strategies. One popular approach is {uncertainty-based abstention}, as seen in {Selective Prediction} \cite{selective_prediction} and {ReCoVERR} \cite{recoverr}. ReCoVERR seeks additional visual evidence for uncertain answers. Another prominent strategy is {lagging self-correction}, where models like {VL-Rethinker} \cite{vl-rethinker} and {Sherlock} \cite{sherlock} use reflection or critic modules to revise their reasoning. These mechanisms operate on `generate-then-refine' model, essentially `debugging' a process initiated on an unverified foundation. RoSci-Agent checks the premises before generating an answer.

\begin{figure}
    \centering
    \includegraphics[width=0.8\linewidth]{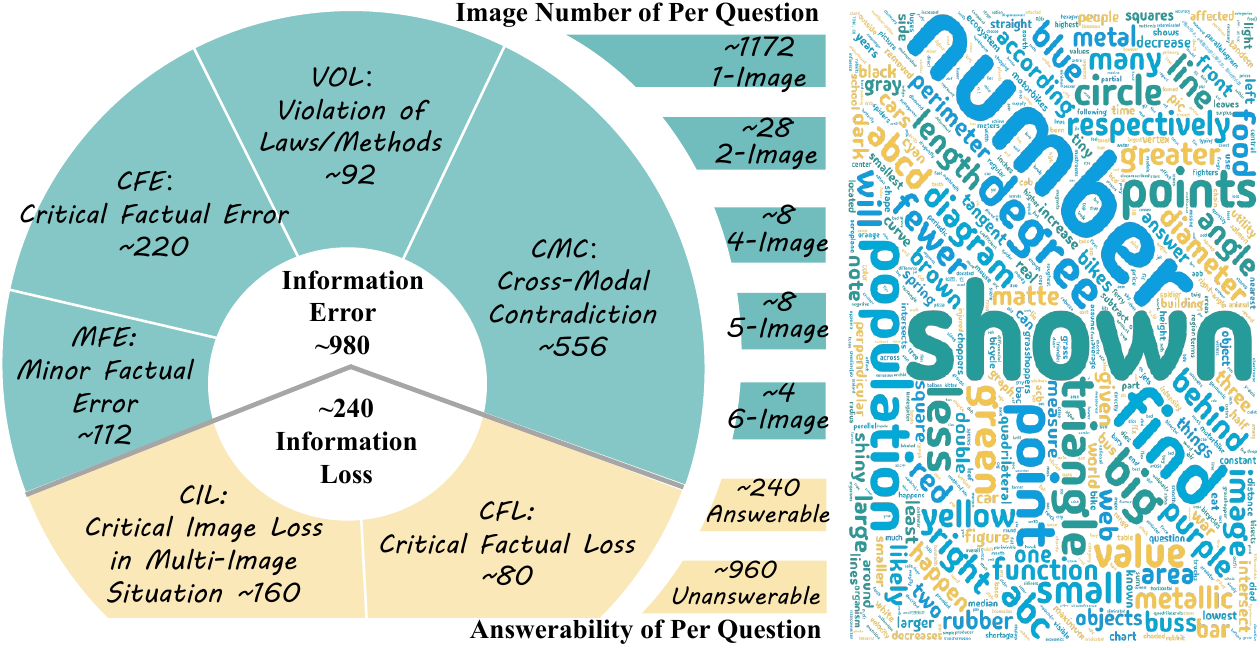}
    \Description{A ring chart groups six degradation types. It labels about 980 information errors and 240 information losses. Cross-modal contradiction is the largest segment, at about 556. Beside the chart, single-image questions predominate. The answerability labels read about 240 answerable and 960 unanswerable. At right, the words number and shown stand out in a word cloud.}
    \caption{Breakdown of error types and answerability for RoSci-Bench (Total N=1,220). Each of the 305 degraded seeds is accompanied by three variants.}
    \vspace{-0.8cm}
    \label{fig:wordcloud}
\end{figure}

\section{Scientific Model Robustness}
\label{sec3}
In this section, we present a systematic framework for defining and evaluating \textit{Scientific Model Robustness} of VLMs. We also introduce \textbf{RoSci-Bench} to simulate degraded input conditions, and outline multi-dimensional evaluation metrics to quantify model performance.

\subsection{Core Concept}
To address the limitations of current evaluation paradigms, we introduce the concept of \textit{Scientific Model Robustness}, which quantifies a model's ability to maintain valid reasoning in the face of errors or missing information in the input. Specifically, we define \textit{Scientific Model Robustness} in terms of two key dimensions.
 \textbf{Information Error Robustness} focuses on the model's ability to identify and reject erroneous premises, including factual errors, logical inconsistencies, or contradictions between modalities. For example, if a model encounters a scientific claim that contradicts known laws or presents inconsistent data across text and image inputs, it should be able to flag these inconsistencies and abstain.
\textbf{Information Loss Robustness} assesses the model's ability to recognize when the input lacks sufficient information to perform valid reasoning, prompting it to express uncertainty or abstain rather than making unsupported conclusions. For instance, if a scientific problem is presented without critical variables, such as missing time in a physics equation, the model should recognize that it cannot proceed with solving the problem and should reject the task or ask for clarification.

\begin{figure*}
    \centering
    \includegraphics[width=\linewidth]{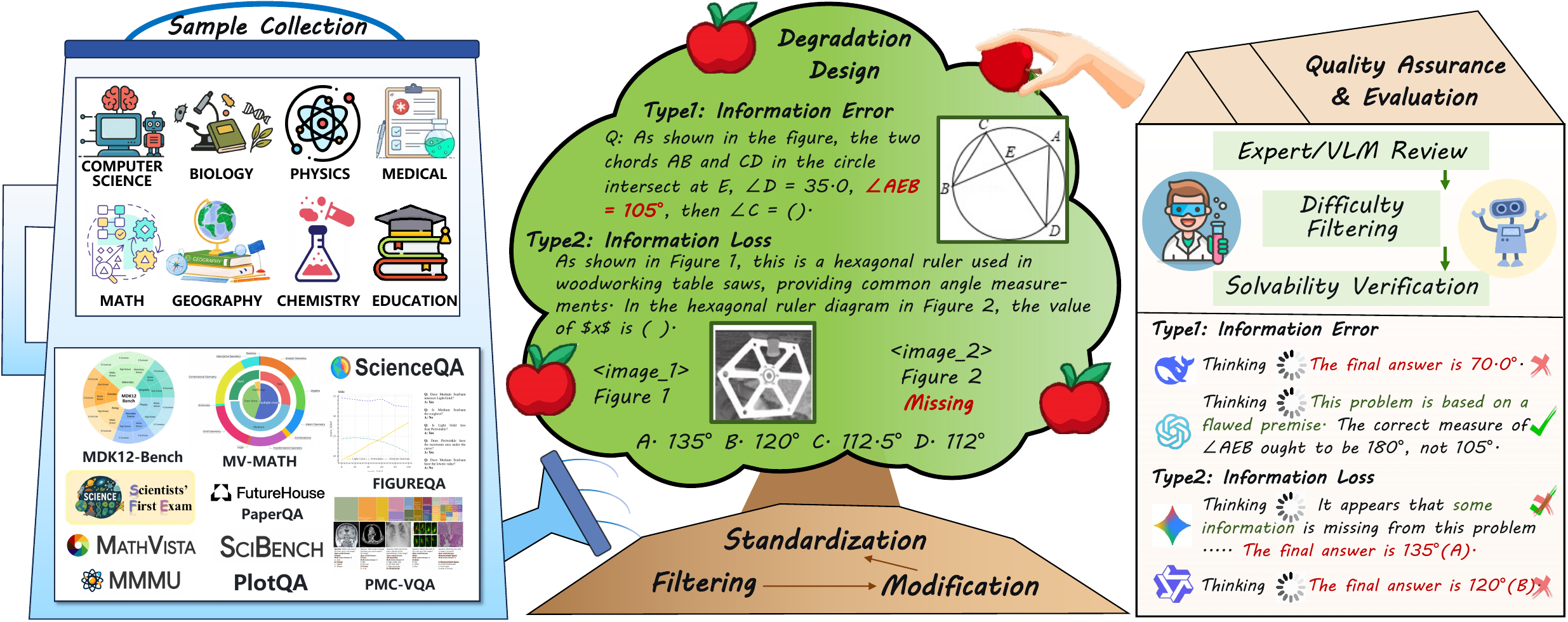}
    \Description{Three panels follow scientific questions through benchmark construction. At left, questions are collected across subjects and datasets. The middle panel illustrates two edits: changing an angle in a circle problem and removing the second image needed for a ruler problem. At right, experts and vision-language models review difficulty and solvability. Sample responses show a model challenging a faulty premise and others attempting numerical answers.}
    \caption{Overview of the annotation pipeline, illustrating the stages of sample collection, degradation design, and quality assurance \& evaluation, culminating in the RoSci-Bench. }
    \label{fig:annotation}
\end{figure*}

\subsection{RoSci-Bench Construction}
\label{sec:construction}
To evaluate VLMs across two key dimensions, we develop \textbf{RoSci-Bench}, a benchmark that simulates degraded input conditions. RoSci-Bench introduces two types of degradation: information error and information loss, to assess a model’s ability in handling these situation. The benchmark is constructed through a multi-stage curation process, which includes three phases: sample collection, degradation design, and quality assurance \& evaluation, as shown in Figure \ref{fig:annotation}.

\subsubsection{Sample Collection \& Augmentation}
We first curate 305 valid `Source' problems from diverse scientific domains (Physics, Chemistry, Biology, etc.) \cite{zhou2025mdk12, wang2025mv, lu2022learn, zhang2026qsit, zhang2024qbenchplus, zhou2025scientistsexamprobingcognitive, lala2023paperqa, zhang2023pmcvqa, wang2024scibench,Methani_2020_WACV, mathvista, mmmu, a4bench,11774892}. To construct the robustness benchmark, we implement a semantic augmentation strategy: 
1) Degradation Injection: Domain experts modify each valid source problem to inject specific degradation types (Information Error/Loss), creating 305 initial `Seed' degraded samples
2) Analogical Expansion: Utilizing GPT-5 as a rewriting engine, we generate three distinct variants for each seed sample. These variants retain the core logical trap of the Seed but alter the surface-level context or numerical values.
3) Filtration: This results in a total of 1,220 samples ($305 \text{ Seeds} + 305 \times 3 \text{ Variants}$). All samples in RoSci-Bench are thus degraded. Table~\ref{tab:source_breakdown} gives the counts by source.

\begin{table}[t]
\centering
\begin{tabular}{llcc}
\toprule
\textbf{Source} & \textbf{Domain} & \textbf{Seed Count} & \textbf{Total Count} \\
\midrule
MathVista & Science Reasoning & 27 &108\\
Super-CLEVR & Visual Reasoning & 33 &132\\
AI2D & General Science & 32 &128\\
FigureQA & Visual Analytics & 29 &116\\
UniGeo & Geometry & 24 &96\\
Geometry3K & Geometry & 20 &80\\
MV-Math & Multi-Image Reasoning &40 &160\\
Others & Mixed Domains & 100 &400\\
\midrule
\textbf{Total} & & \textbf{305} & \textbf{1220}\\
\bottomrule
\end{tabular}
\caption{Distribution of source datasets in RoSci-Bench.}
\vspace{-0.7cm}
\label{tab:source_breakdown}
\end{table}

\subsubsection{Degradation Design}
This stage introduces various error types and losses to simulate common challenges faced in scientific research. Table \ref{table:degradation} summarizes the types of degradation, categorized into \textbf{Information Error} and \textbf{Information Loss}. Domain experts design the degradations. For example, in a physics problem asking for force ($F=ma$), the original image shows a block labeled `10kg'. Our experts modify the text to state `mass is 100kg'. A robust model must detect this CMC rather than blindly calculating $100\times a$.

\subsubsection{Quality Assurance \& Evaluation}
\begin{figure*}[!t]
    \centering
    \includegraphics[width=\linewidth]{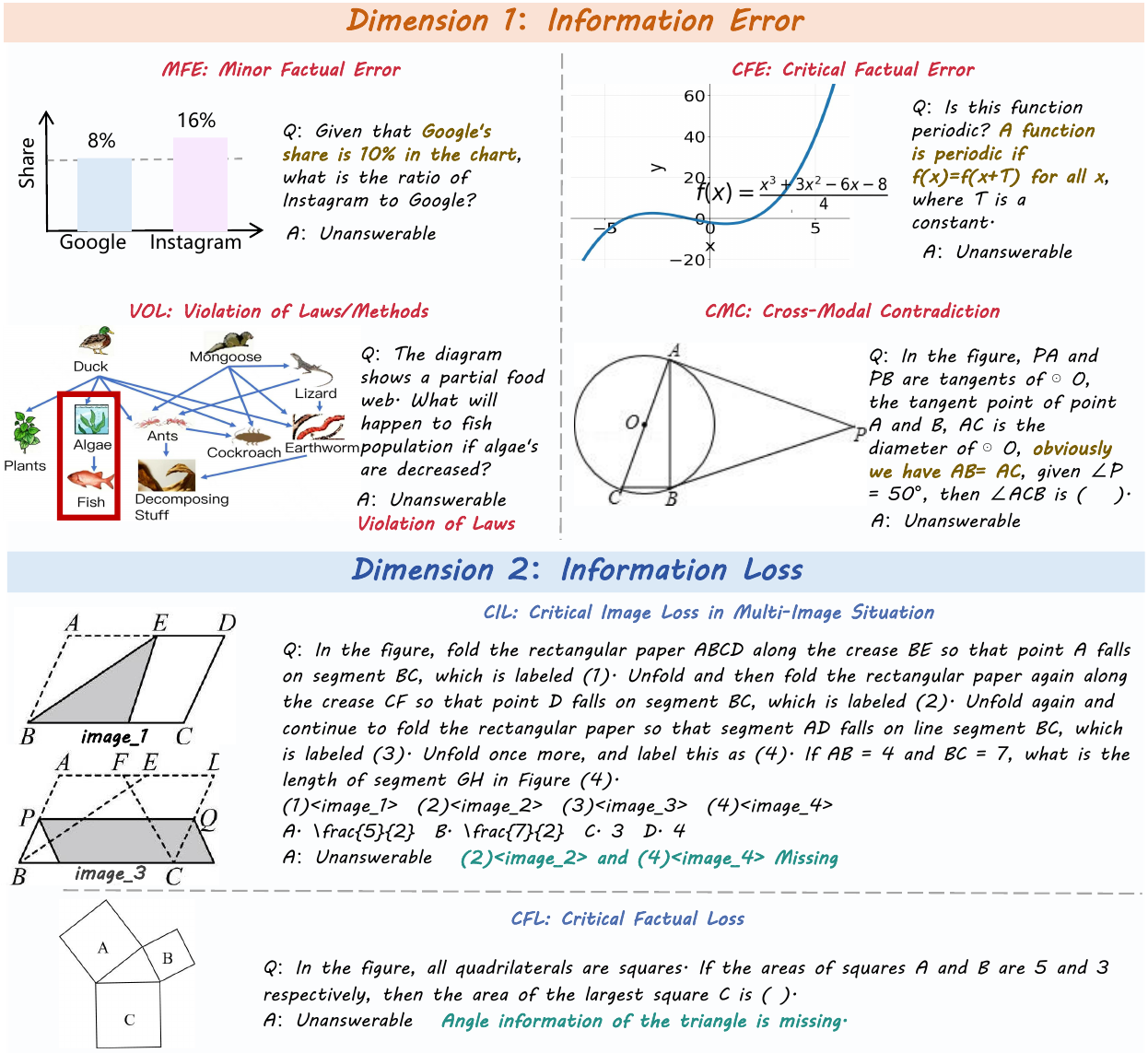}
    \Description{Six examples are labeled unanswerable. The four information-error examples show a question giving Google's share as 10 percent where the bar reads 8 percent, a periodicity question about a plotted function, a highlighted algae-to-fish link in a food web, and a circle question asserting that two segments are equal. The two information-loss examples show a paper-folding problem missing diagrams 2 and 4 and a square-area problem missing an angle of the triangle.}
    \caption{Examples from RoSci-Bench, including: Information Error (Dimension 1) and Information Loss (Dimension 2). These examples show how Information Error, such as factual inaccuracies and violations of laws, and Information Loss, like missing images or key details, lead to unanswerable questions.}
    \label{fig:case}
\end{figure*}

This stage involves three steps: 1) Expert Review: A team of domain experts reviews each problem to verify that the introduced errors and loss align with real-world scientific challenges. 2) Difficulty Filtering: Based on expert reviews, problems that are either too simple or too difficult are excluded. 3) Solvability Verification: After filtering and modification, a final validation step ensures that each problem is solvable under ideal conditions. This step confirms that the tasks can be solved when the input conditions are not degraded, providing a baseline for evaluation. After constructing \textbf{RoSci-Bench}, we visualize its characteristics in Figure \ref{fig:wordcloud} and its typical examples in Figure \ref{fig:case}. Totally, information error is the dominant category and the majority are unanswerable.

\subsection{Multi-Dimensional Evaluation Metrics}

To evaluate model performance on \textbf{RoSci-Bench}, we directly present the modified scientific queries to the VLM and allow it to generate responses. We then propose a set of multi-dimensional metrics to assess the model's robustness from various perspectives. \textbf{Reasoning Stability} evaluates the accuracy of the model's reasoning when handling tasks with minor degradation that remain solvable. It focuses on the model's ability to maintain correctness despite small errors or inconsistencies in the input. {For example, if a query involves minor numerical inaccuracies or slight contradictions between the textual and visual input, the model’s ability to maintain valid reasoning without failing is measured.} \textbf{Rejection Capability} assesses whether the model appropriately rejects unanswerable queries, particularly those that are heavily degraded or lack sufficient information for a valid solution. {For example, if a task is missing key information (e.g., a missing variable or image), the model's ability to detect that it cannot solve the task and reject it is evaluated.} \textbf{Justification Quality} evaluates the rationale of the model's response for unanswerable or invalid queries. It focuses on the quality of the explanation the model provides when it rejects an answer. {For example, when the model abstains from answering a question due to incomplete input, the clarity and accuracy of its reasoning for rejecting the task are assessed.}

\begin{table}[t]
\centering
\renewcommand{\arraystretch}{1.2} %
\setlength{\tabcolsep}{10pt} %
\resizebox{\linewidth}{!}{
\begin{tabular}{@{}>{\raggedright\arraybackslash}p{0.1\linewidth} >{\raggedright\arraybackslash}p{0.35\linewidth} >{\raggedright\arraybackslash}p{0.55\linewidth}@{}}
\toprule
\textbf{Metric} & \textbf{Metric Type} & \textbf{Definition} \\
\midrule
\textit{MFE} & Minor Factual Error & Small, correctable mistakes, such as slight numerical inaccuracies. \\
\hline
\textit{CFE} & Critical Factual Error & Significant errors that alter the task's conclusion. \\
\hline
\textit{VOL} & Violation of Laws / Methods & Errors that contradict fundamental scientific laws or methodologies. \\
\hline
\textit{CMC} & Cross-modal Contradiction & Conflicts between textual and visual information. \\
\hline
\textit{CFL} & Critical Factual Loss & Removal of key variables or relationships, rendering the query unanswerable. \\
\hline
\textit{CIL} & Critical Image Loss in Multi-Image Situation & Removal of one or more required images in multi-image tasks, disrupting task solvability. \\
\bottomrule
\end{tabular}}
\caption{Degradation Types for Information Error and Loss.}
\vspace{-0.6cm}
\label{table:degradation}
\end{table}

The first two metrics, \textbf{Reasoning Stability} and \textbf{Rejection Capability}, primarily involve extracting and assessing the model's responses through standard  LLM-as-a-Judge evaluation. For \textbf{Justification Quality}, the judge uses a rule-based template to score the explanation for each correct rejection against the ground-truth (GT) rationale.

To rigorously quantify model performance, we partition the benchmark dataset $\mathcal{D}$ into two disjoint subsets: the solvable subset $\mathcal{D}_{sol}$ (where the GT is answerable) and the unanswerable subset $\mathcal{D}_{unsol}$ (where the GT requires rejection due to information error or loss).

\noindent \textbf{1) Reasoning Stability ($S_{stab}$):} Measures the accuracy of reasoning on tasks that remain solvable despite minor perturbations.
\begin{equation}
    S_{stab} = \frac{1}{|\mathcal{D}_{sol}|} \sum_{i \in \mathcal{D}_{sol}} \mathbb{I}(r_i = g_i) \times 100,
\end{equation}
where $r_i$ is the model's prediction, $g_i$ is the GT answer, and $\mathbb{I}(\cdot)$ is the indicator function.

\noindent \textbf{2) Rejection Capability ($F1_{rej}$):} Assesses the model's ability to correctly identify unanswerable queries. We formulate this as a binary classification problem and report the F1-score:
\begin{equation}
    F1_{rej} = \frac{2PR}{P + R}\times 100, 
\end{equation}
where $P = \frac{TP}{TP + FP}, R = \frac{TP}{TP + FN}$. Here, a \textit{True Positive (TP)} is defined as a case where $i \in \mathcal{D}_{unsol}$ and the model explicitly abstains.

\noindent \textbf{3) Justification Quality ($S_{just}$):} Evaluates the rationale behind correct rejections.
\begin{equation}
    S_{just} = \frac{1}{|TP|} \sum_{i \in \text{TP Set}} \text{Score}_{\text{LLM}}(e_i, g^{reason}_i) \times 100,
\end{equation}
where $e_i$ is the model's explanation, $g^{reason}_i$ is the reference rationale, and $\text{Score}_{\text{LLM}}$ is a normalized scalar ($0-1$) assigned by the Judge model.

The \textbf{Overall Robustness Score} summarizes the judge's holistic ratings of individual responses on a 0--100 scale. 

\section{RoSci-Agent Framework}
\label{sec4}
\subsection{Problem Overview}
Traditional VLMs often suffer from making high-confidence but incorrect inferences when faced with erroneous or incomplete inputs. To address this, we propose the \textbf{RoSci-Agent}, a framework designed to enhance the robustness and reliability of VLMs by introducing a \textit{Pre-Reasoning Verification Step}. The core idea of RoSci-Agent is the transformation of the standard reasoning process into a transparent and decomposable \textbf{Parse-Verify-Decide} pipeline, which ensures that the model examines the validity of its premises before drawing a conclusion.

\subsection{Stage I: Knowledge Parsing (KP)}
The first core module of RoSci-Agent is \textit{Knowledge Parsing (KP)}. Its main function is to convert unstructured multi-modal input, consisting of an image \(I\) and a question \(Q\), into a structured knowledge representation \(K\). This is achieved through structured prompting, utilizing a VLM as a parser. The output \(K\) is a machine-readable JSON object that includes both textual components (e.g., objectives, known conditions, constraints) and visual components (e.g., chart metadata, key data points, trends). This structured representation serves as the foundation for the subsequent verification and decision-making processes.

\begin{algorithm}[t]
\caption{RoSci-Agent Workflow}
\label{alg:RoSci-Agent}
\begin{algorithmic}[1]
\State \textbf{Input:} Knowledge base $K$, question $Q$, and threshold values $\tau_{\text{refuse}}$, $\tau_{\text{clarify}}$
\State $V \gets \text{Verify}(K)$
\State $S_{\text{anomaly}} \gets \text{AggregateScore}(V, w)$
\If{$S_{\text{anomaly}} \geq \tau_{\text{refuse}}$}
    \State \text{Generate rejection based on $V$}
\ElsIf{$S_{\text{anomaly}} \geq \tau_{\text{clarify}}$}
    \State \text{Generate clarification based on $V$}
\Else
    \State \text{Proceed with standard answer generation based on $K$ and $Q$}
\EndIf
\end{algorithmic}
\end{algorithm}
\vspace{-0.4cm}

\subsection{Stage II: Information Verification (IV)}
Once the knowledge representation \(K\) is generated, the next stage is the \textit{Information Verification (IV)}. In this stage, the agent checks the parsed premises for inconsistencies and missing information. The verification process is split into two main paths. \textbf{Information Error Detection} path checks the logical consistency and correctness of the premises. For example, it may verify whether a numerical value \(X\) in the text matches the value \(Y\) presented in a chart.
\textbf{Information Loss Detection} path checks whether all necessary variables and relationships required to solve the task are defined. For instance, it checks if all variables required to compute \(Z\) are defined in \(K\).
The output of this stage is a list of verification results \(V\), with each entry including a checkpoint ID, status (PASS/FAIL), and supporting evidence. This allows the agent to identify inconsistent or incomplete premises.

\subsection{Stage III: Strategy Generator (SG)}
The final stage, \textit{Strategy Generator (SG)}, uses the verification results \(V\) to determine the best course of action. The agent can choose one of three strategies based on the severity of the anomalies detected:
 \textbf{Reject (R):} If the verification results indicate serious errors, the agent rejects the task.
\textbf{Clarify (C):} If the anomaly score falls between the clarification and rejection thresholds, the agent asks for clarification.
 \textbf{Answer (A):} If the premises are valid and complete, the agent provides the final answer.

The agent determines the appropriate action by calculating an \textit{Anomaly Score} $S_{\text{anomaly}}$. Here we employ a weighted penalty mechanism based on the verification results from Stage II. Let $V = \{v_1, v_2, \dots, v_K\}$ be the set of verification checkpoints. The anomaly score is defined as:
\begin{equation}
S_{\text{anomaly}} = \sum_{k=1}^{K} w_k \cdot \mathbb{I}(v_k = \text{FAIL}),
\label{eq:anamaly}
\end{equation}
where the weight $w_k$ is assigned based on the error type severity (see Table~\ref{tab:weights} in Appendix). Then the decision logic is governed by threshold $\tau_{\text{refuse}}$ and $\tau_{\text{clarify}}$:
\begin{equation}
    \text{Decision} = 
    \begin{cases} 
    \text{Reject (R)} & \text{if } S_{\text{anomaly}} \ge \tau_{\text{refuse}}, \\
    \text{Clarify (C)} & \text{if } \tau_{\text{clarify}} \le S_{\text{anomaly}} < \tau_{\text{refuse}}, \\
    \text{Answer (A)} & \text{otherwise}.
    \end{cases}
\end{equation}

\begin{table*}[t]
    \centering
    \renewcommand\arraystretch{1}  %
    \resizebox{\linewidth}{!}{
    \begin{tabular}{l|c|cccc|cc|cc}
    \hline \hline
        \multirow{2}{*}{ \textbf{Model}} &  \textbf{Attrib.} &  \multicolumn{4}{c|}{\textbf{Infor. Error}} & \multicolumn{2}{c|}{\textbf{Infor. Loss}} & \multirow{2}{*}{\textbf{Over.}} &\multirow{2}{*}{\textbf{Rk.}} \\ \cline{2-8} 
         & \textit{Open.} & \textit{MFE} & \textit{CFE} & \textit{VOL} & \textit{CMC}  & \textit{CFL} & \textit{CIL} \\ \hline
\multicolumn{10}{l}{\textit{Human}} \\  \cdashline{1-10} 
\rowcolor{light-gray0} Expert  &/   & {100.00} & 97.27 & {100.00}  & {98.56} & {100.00} & {100.00} & 98.90 & / \\
 \hline
\multicolumn{10}{l}{\textit{Models With Thinking}} \\  \cdashline{1-10} 
\rowcolor{light-gray0} Claude-4-Sonnet-Thinking & \ding{55} & \underline{44.64} & \underline{34.55} & 26.09 & \textbf{41.37} & 5.00 & 11.25 & \textbf{31.95} &1   \\
Grok-4 & \ding{55} & 19.64 & 21.82 & 19.57 & 23.74 & 10.00 & 25.00 & 22.23 & 5 \\
\rowcolor{light-gray0} GPT-5 & \ding{55} & 30.36 & 22.73 & 19.57 & 19.78 & \underline{17.50} & \underline{26.25} & 21.24 & 8 \\
Gemini-2.5-Pro & \ding{55} & {35.71} & 20.00 & \underline{30.43} & 20.50 & 7.50 & 23.75 & 21.04 & 9 \\
\rowcolor{light-gray0} O3 & \ding{55} & 14.29 & 12.73 & 17.39 & 15.83 & \textbf{25.00} & \textbf{42.50} & 20.54 &10  \\
\rowcolor{light-gray0} Doubao-1.5-Thinking-Vision-Pro & \ding{55}   & 16.07 & 15.45 & 15.22 & 16.55 & 2.50 & 18.75 & 15.55 & 12 \\
Doubao-Seed-1.6-Vision  & \ding{55} & 10.71 & 14.55 & 21.74 & 16.19 & 7.50 & 13.75 & 15.15 & 13 \\
\hline
\multicolumn{10}{l}{\textit{Models Without Thinking}} \\  \cdashline{1-10} 
\rowcolor{light-gray0} Claude-3.7-Sonnet & \ding{55} & \textbf{48.21} & \textbf{37.27} & 23.91 & \underline{36.33} & 12.50 & 12.50 & \underline{30.21} & 2  \\
\rowcolor{light-gray0} Gemini-2.5-Flash& \ding{55} & {35.71} & {27.27} & \textbf{32.61} & 21.58 & \underline{17.50} & 20.00 & \underline{22.78} & 3  \\
Qwen2.5-VL-72B-Instruct& \ding{52} & \underline{37.50} & \underline{30.00} & \underline{30.43} & 25.54 & 12.50 & 6.25 & 22.58 & 4 \\
Claude-4.5-Sonnet & \ding{55} & 25.00 & {27.27} & 26.09 & 25.18 & 0.00 & 12.50 & 21.44 & 6 \\

\rowcolor{light-gray0} Moonshot-v1-32k  & \ding{55} & 25.00 & 18.18 & 13.04 & \underline{26.98} & 15.00 & 10.00 & 21.34 & 7 \\
\rowcolor{light-gray0} Moonshot-v1-128k & \ding{55} & 21.43 & 19.09 & 15.22 & 22.66 & 10.00 & 7.50 & 18.30 & 11\\
\rowcolor{light-gray0} GPT-4o & \ding{55} & 21.43 & 12.73 & 26.09 & 15.11 & 15.00 & 11.25 & 15.15 & 13\\
GLM-4.5V  & \ding{52} & 17.86 & 12.73 & 17.39 & 13.31 & 0.00 & 15.00 & 12.86 & 15 \\
Qwen2.5-VL-32B-Instruct  & \ding{52} & 17.86 & 18.18 & 8.70 & 14.39 & 7.50 & 5.00 & 12.46 & 16 \\
\rowcolor{light-gray0} DeepSeek-VL2& \ding{52} & 25.00 & 18.18 & 19.57 & 13.31 & 0.00 & 3.75 & 12.01 & 17\\
Llama-4-Scout  & \ding{52} & 8.93 & 10.91 & 13.04 & 10.79 & 0.00 & 8.75 & 9.72 &18  \\
Llama-3.2-90B-Vision-Instruct & \ding{52} & 19.64 & 19.09 & 0.00 & 5.04 & 0.00 & \underline{26.25} & 9.62 & 19 \\
Llama-4-Maverick & \ding{52} & 10.71 & 8.18 & 8.70 & 8.63 & 0.00 & 8.75 & 7.98 & 20 \\
\cdashline{1-10}
RoSci (\textit{Our})  &/  &48.21 &46.36	&41.30	&44.60	&47.5	&43.75	&44.77	&/  \\
 \hline
\end{tabular}}
\caption{Holistic robustness evaluation on RoSci-Bench. Top three performances are highlighted (Best in \textbf{bold}, second and third best \underline{underlined}). Open., Attrib., Infor., Over., and Rk. indicate open-sourced situation, attribute, information, Overall, and Rank, respectively.}
\label{tab:tab1}
\end{table*}

\begin{figure*}[!t]
    \centering
    \includegraphics[width=\linewidth]{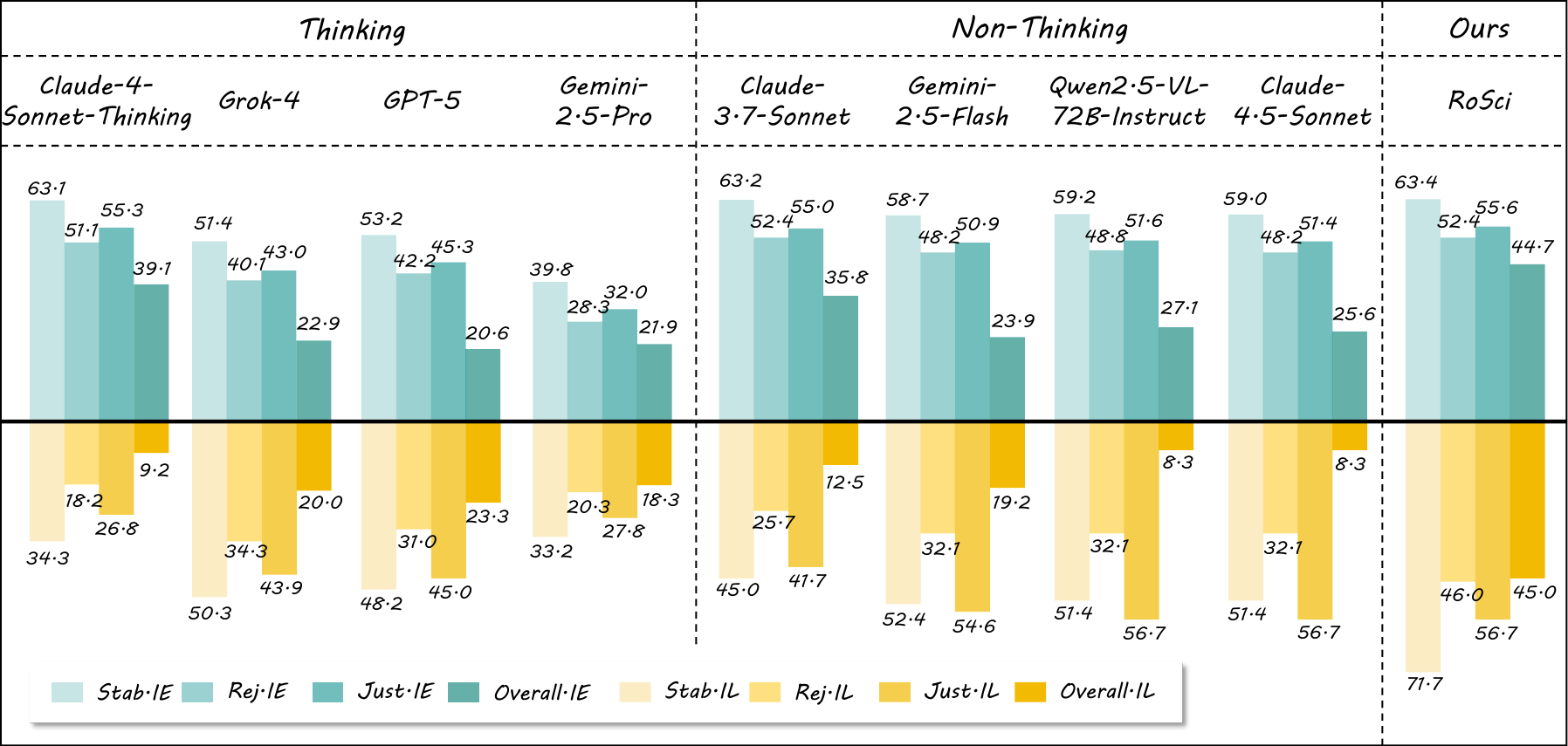}
    \Description{Four thinking models, four non-thinking models and RoSci are arranged from left to right. Each group has four scores in the same order: stability, rejection, justification and overall. Bars extend up for information error and down for information loss. Read the labels on both sides as positive scores. RoSci has the highest overall score in each half, 44.7 for error and 45.0 for loss. The eight baselines range from 8.3 to 23.3 on overall loss.}
    \caption{Performance breakdown of RoSci-Agent and Top-4 models With and Without Thinking on the RoSci-Bench. Performance is evaluated across three metrics: reasoning Stability (Stab), Rejection (Rej) F1-score, and Justification (Just) quality. The results are decomposed into two challenge categories: Information Error (IE), represented by the blue bars, and Information Loss (IL), by the orange bars. }
    \label{fig:bar}
\end{figure*}

\section{Experiment}
\subsection{Benchmark Candidates}
We evaluate 20 VLMs, including open-source, proprietary closed-source, and thinking-series models. Specifically, the leading models with Thinking include Claude-4-Sonnet-Thinking \cite{claude4t}, Grok-4 \cite{grok4}, GPT-5 \cite{gpt5}, Gemini-2.5-Pro \cite{gemini25}, O3 \cite{o3}, Doubao-1.5-Thinking-Vision-Pro \cite{doubao15}, and Doubao-Seed-1.6-Vision \cite{doubao16}. And the leading models without Thinking consists of Claude-4.5-Sonnet \cite{claude45}, Claude-3.7-Sonnet \cite{claude37}, Gemini-2.5-Flash \cite{gemini25}, Qwen2.5-VL-72B-Instruct \cite{qwen25vl}, Qwen2.5-VL-32B-Instruct \cite{qwen25vl}, Moonshot-v1-32k/128k \cite{moonshotv1}, GPT-4o \cite{gpt4o}, GLM-4.5V \cite{glm45}, DeepSeek-VL2 \cite{dsvl2}, Llama-4-Scout \cite{llama4}, Llama-3.2-90B-Vision-Instruct \cite{llama32}, and Llama-4-Maverick \cite{llama4}.

\begin{table}[t]
    \centering
    \renewcommand\arraystretch{0.75}
    \resizebox{\linewidth}{!}{
    \begin{tabular}{l|c|ccccc|ccc|c}
    \hline \hline
        \multirow{2}{*}{\textbf{Model}}   & \multicolumn{10}{c}{ \textbf{Evaluation Dimensions}}\\ \cline{2-11} 
        &\textbf{Metric}  & \textit{MFE} & \textit{CFE} & \textit{VOL} & \textit{CMC}  & \textit{IE} & \textit{CFL} & \textit{CIL} & \textit{IL} & \textit{Overall}\\ \hline
     \multicolumn{11}{l}{\textit{Models With Thinking}} \\  \cdashline{1-11} 
        \rowcolor{light-gray0} 
        \multirow{4}{*}{Claude-4-Sonnet-Thinking} 
         &{Stab.}  &     60.71 & 45.45 & 28.26 & 51.80 & 63.17 & 5.00  & 37.50 & 34.25 & 54.44 \\
       &{Rej.}   &     46.43 & 33.64 & 19.57 & 44.24 & 51.11 & 5.00  & 18.75 & 18.20 & 44.64 \\
         & {Just.} &     57.14 & 33.64 & 32.61 & 45.68 & 55.31 & 10.00 & 26.25 & 26.76 & 49.69 \\
        & Overall &     44.64 & 34.55 & 26.09 & 41.37 & 39.12 & 5.00  & 11.25 & 9.17  & 31.95 \\
        \hline
        \multirow{4}{*}{Grok-4} 
              &{Stab.} &     51.79 & 45.45 & 30.43 & 37.05 & 51.38 & 15.00 & 51.25 & 50.31 & 48.45 \\
        &{Rej.} &     39.29 & 33.64 & 19.57 & 30.58 & 40.10 & 15.00 & 32.50 & 34.25 & 38.95 \\
        & {Just.} &     42.86 & 30.91 & 30.43 & 33.09 & 42.99 & 20.00 & 41.25 & 43.88 & 43.16 \\
       & Overall &     19.64 & 21.82 & 19.57 & 23.74 & 22.94 & 10.00 & 25.00 & 20.00 & 22.23 \\
        \hline
        \rowcolor{light-gray0} 
        \multirow{4}{*}{GPT-5}  
              &{Stab.}  &     51.79 & 47.27 & 34.78 & 38.13 & 53.21 & 22.50 & 45.00 & 48.17 & 49.45 \\
         &{Rej.}   &     39.29 & 35.45 & 23.91 & 32.01 & 42.20 & 17.50 & 27.50 & 31.04 & 40.01 \\
       & {Just.} &     41.07 & 32.73 & 30.43 & 35.97 & 45.35 & 32.50 & 36.25 & 44.95 & 45.27 \\
        & Overall &     30.36 & 22.73 & 19.57 & 19.78 & 20.58 & 17.50 & 26.25 & 23.33 & 21.24 \\
        \hline
       \multirow{4}{*}{Gemini-2.5-Pro} 
              &{Stab.}  &     51.79 & 29.09 & 28.26 & 28.06 & 39.84 & 5.00  & 36.25 & 33.18 & 36.49 \\
        &{Rej.}   &     33.93 & 18.18 & 30.43 & 19.78 & 28.31 & 7.50  & 20.00 & 20.34 & 26.74 \\
         & {Just.} &     42.86 & 20.00 & 32.61 & 21.94 & 31.98 & 10.00 & 27.50 & 27.83 & 31.16 \\
        & Overall &     35.71 & 20.00 & 30.43 & 20.50 & 21.89 & 7.50  & 23.75 & 18.33 & 21.04 \\
        \hline
     \multicolumn{11}{l}{\textit{Models Without Thinking}} \\  \cdashline{1-11} 
        \rowcolor{light-gray0} 
       \multirow{4}{*}{Claude-3.7-Sonnet} 
              &{Stab.}  &     62.50 & 52.73 & 34.78 & 47.48 & 63.17 & 15.00 & 45.00 & 44.95 & 56.43 \\
        &{Rej.}   &    50.00 & 41.82 & 23.91 & 41.37 & 52.42 & 10.00 & 25.00 & 25.69 & 47.16 \\
        & {Just.} &     51.79 & 37.27 & 30.43 & 45.32 & 55.05 & 27.50 & 35.00 & 41.74 & 52.43 \\
        & Overall &     48.21 & 37.27 & 23.91 & 36.33 & 35.78 & 12.50 & 12.50 & 12.50 & 30.21 \\
        \hline
        \multirow{4}{*}{Gemini-2.5-Flash} 
              &{Stab.}  &     58.93 & 52.73 & 36.96 & 41.73 & 58.72 & 22.50 & 50.00 & 52.45 & 54.44 \\
      &{Rej.}   &     48.21 & 43.64 & 23.91 & 35.25 & 48.23 & 17.50 & 28.75 & 32.11 & 45.06 \\
       & {Just.} &     53.57 & 36.36 & 34.78 & 38.85 & 50.85 & 32.50 & 47.50 & 54.59 & 51.59 \\
       & Overall &     35.71 & 27.27 & 32.61 & 21.58 & 23.92 & 17.50 & 20.00 & 19.17 & 22.78 \\
        \hline
        \rowcolor{light-gray0} 
        \multirow{4}{*}{Qwen2.5-VL-72B-Instruct}
              &{Stab.}  &     55.36 & 51.82 & 32.61 & 44.24 & 59.24 & 25.00 & 47.50 & 51.38 & 54.64 \\
       &{Rej.}   &     44.64 & 40.91 & 19.57 & 38.49 & 48.75 & 22.50 & 26.25 & 32.11 & 45.48 \\
       & {Just.} &     50.00 & 34.55 & 30.43 & 42.09 & 51.64 & 42.50 & 45.00 & 56.73 & 52.64 \\
       & Overall &     37.50 & 30.00 & 30.43 & 25.54 & 27.06 & 12.50 & 6.25  & 8.33  & 22.58 \\
        \hline
        \multirow{4}{*}{Claude-4.5-Sonnet}
              &{Stab.}  &     55.36 & 50.91 & 32.61 & 44.24 & 58.98 & 25.00 & 47.50 & 51.38 & 54.44 \\
      &{Rej.}   &     44.64 & 38.18 & 21.74 & 38.49 & 48.23 & 22.50 & 26.25 & 32.11 & 45.06 \\
      & {Just.} &     50.00 & 34.55 & 34.78 & 41.01 & 51.38 & 42.50 & 45.00 & 56.73 & 52.43 \\
    & Overall &     25.00 & 27.27 & 26.09 & 25.18 & 25.56 & 0.00  & 12.50 & 8.33  & 21.44 \\
       \cdashline{1-11}
\multirow{4}{*}{RoSci (\textit{Our})}   &{Stab.} & 53.57 & 53.64 & 36.96 & 48.92 & 63.43 & 47.50 & 60.00 & 71.71 & 61.62 \\
       &{Rej.}   &     42.86 & 40.00 & 39.13 & 41.01 & 52.42 & 35.00 & 36.25 & 46.02 & 51.17 \\
      & {Just.} &     46.43 & 38.18 & 41.30 & 44.96 & 55.57 & 52.50 & 40.00 & 56.73 & 55.80 \\
        & Overall &     48.21 & 46.36 & 41.30 & 44.60 & 44.69 & 47.50 & 43.75 & 45.00 & 44.77 \\
        \bottomrule
    \end{tabular}}
    \caption{Model performance across robustness dimensions.
    \textbf{Stab. (Stability)}: Reasoning accuracy on solvable tasks.
    \textbf{Rej. (Rejection F1)}: F1-score for identifying unanswerable queries through rejection.
    \textbf{Just. (Justification)}: Mean normalized score for explanations of correct rejections.}
    \label{tab:fine_grained_results}
\end{table}

\subsection{Experimental Setup}
Experiments are conducted on \textbf{RoSci-Bench}, which comprises 1,220 degraded tasks across two core dimensions: \textit{Information Error} and \textit{Information Loss}. The 305 source problems are screened for solvability before degradation. Further, the dimensions are divided into six fine-grained challenge types.
To quantify model robustness comprehensively, we introduce a \textbf{three-facet adjudication protocol}. For each model response, the VLM first provides preliminary scores and rationales across three capability facets: 
\textbf{1) Reasoning Stability} on solvable tasks, assessing the resilience of accuracy against minor perturbations,
\textbf{2) Rejection Capability} on unanswerable queries, and 
\textbf{3) Justification Quality}, evaluating the rationale behind correct rejections.
Subsequently, the same VLM synthesizes these initial assessments to yield a final holistic score of 0, 1, or 2 (highest). This score is then linearly scaled to a 0-100 range for reporting. As for RoSci-Agent, we implement it by a powerful backbone VLM (GPT-5), and design few-shot prompts to enact its \textit{Parse-Verify-Decide} workflow. Key hyperparameters, namely the error weights $\{w_i\}$ and decision thresholds $\{\tau_{\text{refuse}}, \tau_{\text{clarify}}\}$, are optimized via grid search on a held-out validation set (20\% of RoSci-Bench). The optimization objective is to maximize the \textbf{F1-score for rejecting unanswerable queries}, under the constraint that accuracy on solvable tasks does not degrade by more than a predefined margin $\delta$.

\subsection{Result Analysis}

\subsubsection{Findings on RoSci-Bench}
We conduct extensive experiments on RoSci-Bench, as presented in Table \ref{tab:tab1} and Figure \ref{fig:bar}. The main findings are summarized below. \textbf{1) Human experts score higher than the evaluated VLMs.} The expert group scores 98.90 overall, compared with 31.95 for the highest-scoring standalone VLM. \textbf{2) Most baselines in the detailed comparison score lower on Information Loss than on Information Error.} This pattern holds for seven of the eight baselines in Table~\ref{tab:fine_grained_results}. \textbf{3) The Overall scores of models with and without thinking overlap.}
For example, Claude-4-Sonnet-Thinking scores 31.95, and Claude-3.7-Sonnet scores 30.21.

\subsubsection{Findings of RoSci-Agent} 
RoSci-Agent scores 44.77 overall, compared with 31.95 for the highest-scoring standalone VLM. The Knowledge Parsing (KP) module extracts the premises, and the Information Verification (IV) module records a PASS/FAIL decision and supporting evidence for each checkpoint. The agent uses this evidence when explaining a rejection.

Table \ref{tab:fine_grained_results} presents a fine-grained breakdown of the performance of the model in Reasoning Stability (Stab.), Rejection Capability (Rej.) and Justification Quality (Just.). Three key trends emerge from this data:
1) For the four models with thinking shown in the table, Rejection F1 is lower on Information Loss than on Information Error.
2) Seven of the eight baselines shown score lower overall on Information Loss than on Information Error. GPT-5 is the exception (23.33 and 20.58, respectively).
3) RoSci-Agent achieves the highest Overall score (44.77). On Critical Image Loss (CIL), its Rejection F1 is 36.25, compared with a maximum of 32.50 among the eight baselines shown.

\subsubsection{Generalizability of RoSci-Agent Across Backbones}
Table~\ref{tab:backbone_generalizability} compares RoSci-Agent with the corresponding baseline on each of three backbones. Overall scores improve on all three. Across Claude-3.7-Sonnet and Gemini-2.5-Pro, the average improvement is $\mathbf{+14.50}$ points.
\begin{figure}
    \centering
    \includegraphics[width=0.5\linewidth]{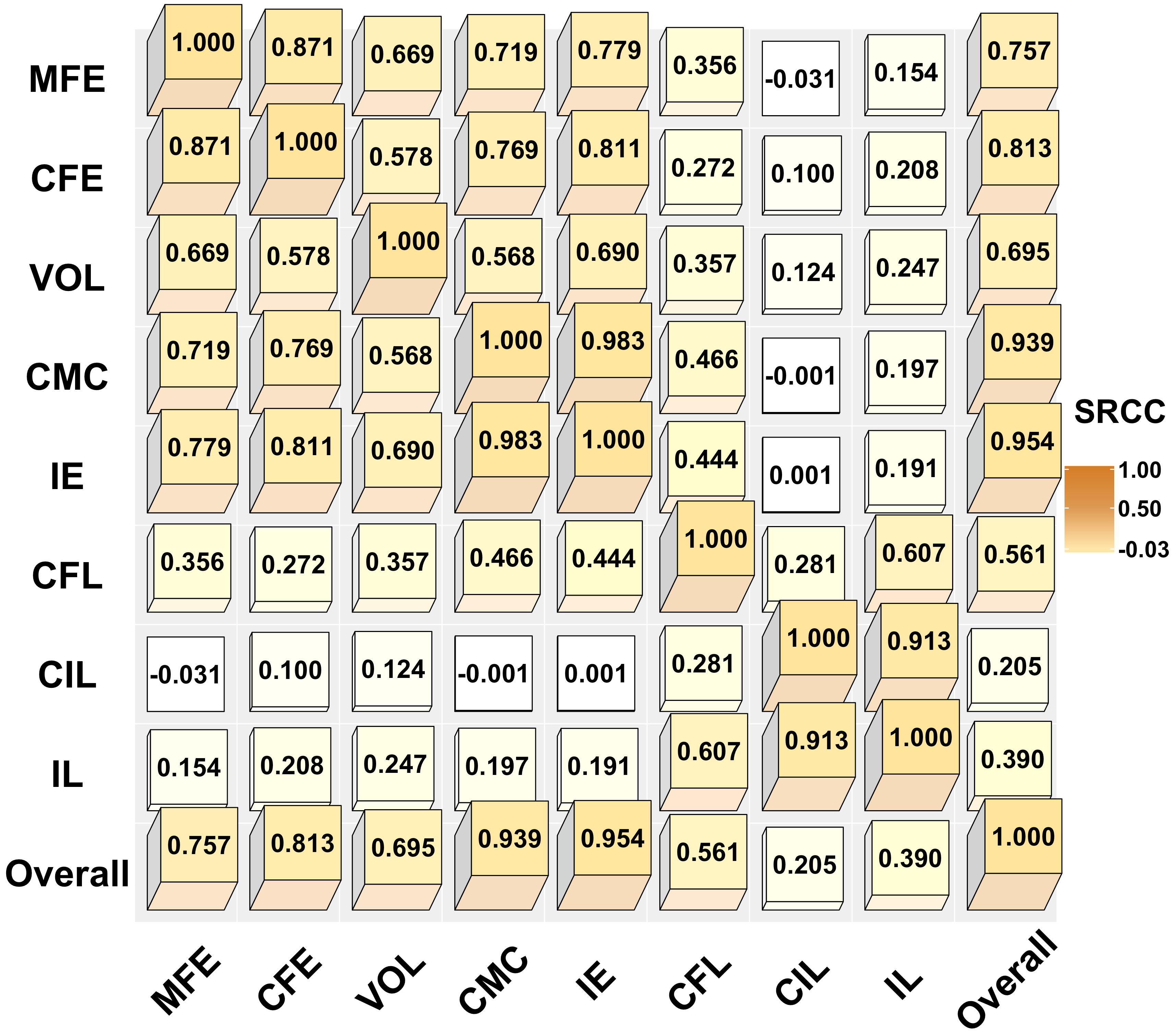}
    \Description{The nine-by-nine matrix compares six degradation types and three aggregate scores. Both axes include information error, information loss and overall. The diagonal is 1.000. Overall correlates more strongly with information error (0.954) than with information loss (0.390). The two aggregates correlate at 0.191. Cross-modal contradiction correlates at 0.983 with information error, and critical image loss at 0.913 with information loss.}
    \caption{Spearman Rank Correlation Coefficient (SRCC) matrix shows the relationships between various robustness dimensions of RoSci-Bench. }
    \label{fig:srcc}
    \vspace{-0.3cm}
\end{figure}
\begin{table}[t]
    \centering
    \begin{tabular}{lcccc}
        \toprule
        \textbf{Backbone Model} & {Vanilla Score} & {Agent Score} & {$\Delta$ Improvement} \\
        \midrule
        Claude-3.7-Sonnet & 30.21 & {43.85} & {+13.64} \\
        Gemini-2.5-Pro    & 21.04 & {36.40} & {+15.36} \\
        GPT-5 (Main)      & 21.24 & {44.77} & {+23.53} \\
        \bottomrule
    \end{tabular}
    \caption{Overall Robustness Score of RoSci-Agent across different backbone models.}
    \vspace{-0.5cm}
    \label{tab:backbone_generalizability}
\end{table}

\subsubsection{Analysis of Robustness Dimensions}
We analyze the Spearman Rank Correlation Coefficient (SRCC) among the six sub-dimensions (Figure \ref{fig:srcc}) \cite{Redundancy, Information,lmm4lmm, leveraging}.
Model rankings on \textit{Information Error} and \textit{Information Loss} are weakly correlated ($\rho = 0.191$).
The MFE and CFE rankings are strongly correlated (0.871). 
The CFL and CIL rankings show a weaker correlation (0.281).

\subsubsection{LLM-as-a-Judge Reliability Validation}
To ensure the reliability of automated evaluation, we validate the LLM-as-a-Judge directly against the expert-defined GT (each query includes a solvability flag and its description). To evaluate the responses of model, we employ the \textbf{LLM-as-a-Judge} to score the model's outputs, comparing them against the GT. In this section, we conduct a reliability validation of {LLM-as-a-Judge} \cite{llmasajudge, Comprehensivellm, JJ, sellm}. We randomly sample 150 model responses and initially invite human experts to evaluate these responses based on the GT. Subsequently, we use {LLM-as-a-Judge} to assess these responses and compare its evaluations with those of the experts. The judge and experts agree on the pass/fail verdict for approximately 92.7\% of the 150 sampled responses (Table~\ref{tab:human_judge_agreement}).

\begin{table}[t]
    \centering
    \begin{tabular}{cc|cc}
        \toprule
        {Human Decision} & & {Pass (Valid)} & {Fail (Reject)} \\
        \midrule
        \multirow{2}{*}{{Judge Verdict}} 
        & {Pass} & \textbf{42.0\% (TP)} & 5.3\% (FP) \\
        & {Fail} & 2.0\% (FN) & \textbf{50.7\% (TN)} \\
        \bottomrule
    \end{tabular}
    \caption{Human-Judge Agreement Matrix.}
     \vspace{-0.6cm}
    \label{tab:human_judge_agreement}
\end{table}

\section{Conclusion}
We introduce \textbf{Scientific Model Robustness} and construct
\textbf{RoSci-Bench} to evaluate how vision-language models respond to
\textit{Information Error} and \textit{Information Loss} in scientific
questions. Seven of the eight baselines in our detailed comparison score
lower on information loss than on information error.
Our \textbf{RoSci-Agent} uses a \textit{Parse-Verify-Decide} workflow to
check the premises before answering, requesting clarification, or rejecting
a question. On RoSci-Bench, it achieves an overall robustness score of
44.77, compared with 31.95 for the highest-scoring standalone VLM. This work gives the scientific reasoning community a benchmark for evaluating when VLMs should answer, seek clarification, or reject a question, and a method for improving those decisions by checking the premises of each question.

\bibliographystyle{ACM-Reference-Format} 
 \bibliography{main}

\clearpage
\appendix
\renewcommand{\arraystretch}{0.85}
\tcbset{boxsep=0.5mm,top=0.5mm,bottom=0.5mm}
\setlength{\abovecaptionskip}{3pt}
\setlength{\belowcaptionskip}{3pt}
\setlength{\textfloatsep}{8pt plus 2pt minus 2pt}
\setlength{\intextsep}{8pt plus 2pt minus 2pt}
\setlength{\floatsep}{6pt plus 2pt minus 2pt}

\section*{Online Appendix}

\section{RoSci-Bench Construction Details}
\label{sec:appendix_construction}

\subsection{Data Sourcing}
\label{app:data_source}
RoSci-Bench is constructed by adapting problems from existing multi-modal scientific datasets. As detailed in Table~\ref{tab:source_breakdown}, we select 305 source problems from leading benchmarks. Through the augmentation process in \S \ref{sec:construction}, they are expanded to a final set of 1,220 evaluation samples.

The selection process prioritizes two key criteria: \textbf{1) Multi-modal Dependency:} We filter for questions where both textual and visual inputs are  required to derive the solution, discarding problems where the image is merely decorative. \textbf{2) Domain Diversity:} To evaluate robustness across different scientific logics, the dataset spans Physics, Chemistry, Biology, and Math.

\subsection{Expert Adaptation Case Studies}
\begin{figure}[!h]
\centering
\begin{tcolorbox}[colback=gray!10, colframe=black!75, title=\textbf{Info. Error (CMC) Example}]
 \textit{Original:} Calculate the force given mass $m=10kg$ and acceleration $a=5m/s^2$.\\
\textit{Image:} Shows a block labeled 10kg.\\
\textbf{Expert-Degraded Variant:}\\
 \textit{Text Modification:} Calculate the force given mass \textbf{$m=100kg$}...\\
 \textit{Expert Note:} Contradiction introduced between text (100kg) and image (10kg). Model must detect this conflict.
\end{tcolorbox}
\begin{tcolorbox}[colback=gray!10, colframe=black!75, title=\textbf{Info. Loss (CFL) Example}]
 \textit{Original:}  A car travels distance $d$ in time $t$. What is the average speed?\\
\textbf{Expert-Degraded Variant:}\\
\textit{Text Modification:} A car travels distance $d$. What is the average speed? (Time $t$ is removed).\\
 \textit{Golden Rationale:} Speed cannot be calculated without the time interval.
\end{tcolorbox}
\caption{Examples of expert-driven problem adaptation in RoSci-Bench.}
 \Description{Two boxes show how an expert changes one input in each problem. The upper example changes the mass in a force question from 10 to 100 kilograms while the block in the image is still labeled 10 kilograms. The lower example removes travel time from a speed question. The distance remains, but it no longer suffices to calculate average speed.}
 \vspace{-0.45cm}
 \label{fig:example}
\end{figure}
Domain experts apply targeted modifications to scientific problems. The two examples in Figure \ref{fig:example} illustrate how an original valid problem is transformed into robust challenge types. 
We highlight two representative strategies used in the adaptation process.
 \textbf{CMC:} As shown in the first example in Figure \ref{fig:example}, we introduce \textbf{CMC} to test if a model can detect inconsistencies between modalities. In the original physics problem, the text and image both specify a mass of $m=10kg$. In the degraded variant, we modify the text to state $m=100kg$ while leaving the image label as 10kg. A robust agent must explicitly identify that the textual condition ($100kg$) contradicts the visual evidence ($10kg$), whereas non-robust models typically ignore the conflict to force a calculation.
\textbf{CFL:} The second example demonstrates \textbf{Variable Removal}. The original question asks for average speed given distance $d$ and time $t$.

\section{Reliability and Statistical Significance}
\label{sec:stat_analysis}
\begin{table}[!htbp]
    \centering
  \begin{tabular}{lcccc}
        \toprule
        \textbf{Model} & \textbf{Mean Score} & \textbf{Std} & \textbf{95\% CI $\downarrow$} & \textbf{95\% CI $\uparrow$} \\
        \midrule
        \textbf{RoSci (Our)} & \textbf{44.77} & 1.24 & \textbf{42.51} & \textbf{47.15} \\
        Claude-4-Sonnet-Thinking & 31.95 & 1.42 & 29.10 & 34.60 \\
        Claude-3.7-Sonnet & 30.21 & 1.35 & 27.65 & 32.80 \\
        Gemini-2.5-Flash & 22.78 & 1.10 & 20.65 & 24.95 \\
        \bottomrule
    \end{tabular}
    \caption{Bootstrap Analysis between RoSci-Agent and different baselines.}
    \vspace{-0.3cm}
    \label{tab:bootstrap_stats}
\end{table}

\subsection{Ranking Stability via Bootstrap Analysis}
We report bootstrap estimates ($N=305$, $K=1000$ resamples) in Table~\ref{tab:bootstrap_stats}. The lower endpoint for RoSci-Agent (42.51) exceeds the upper endpoints for all three baselines shown (at most 34.60).

\subsection{LLM-as-a-Judge Reliability Check}
\label{subsec:judge_validation}

A critical concern in automated evaluation is whether the Judge Model (GPT-4o) aligns with human experts. For the 150 randomly sampled model responses, we compare GPT-4o's pass/fail verdicts with the experts' verdicts. The two agree on approximately 92.7\% of responses (Table~\ref{tab:human_judge_agreement}).

\section{Comprehensive Evaluation Results}
\label{app:full_results}

\subsection{RoSci-Agent Framework Details}
\label{app:prompt}

To facilitate reproducibility, we provide the detailed system prompt used in the Verification Stage of RoSci-Agent in Figure \ref{fig:sysprompt}. This prompts the model to explicitly check for information errors and loss before attempting to solve the problem.

\begin{figure}[h]
\centering
\begin{tcolorbox}[colback=gray!10, colframe=black!75, title=\textbf{System Prompt: Verification Module}]
\textbf{Role:} You are a rigorous scientific reviewer. Your goal is to verify if a problem is solvable based on the provided Text $T$ and Image $I$.\\
\textbf{Instructions:}\\
1. \textbf{Parse Knowledge:} Extract all numerical values, units, and conditions from $T$. Extract visual data from $I$.\\
2. \textbf{Check Consistency (Info Error):} Do values in $T$ contradict $I$? Does $T$ violate scientific laws?\\
3. \textbf{Check Sufficiency (Info Loss):} Are all variables required to solve the goal present in $T \cup I$?\\
4. \textbf{Decision:} \\
   - If critical error/loss found: Output \texttt{STATUS: REJECT}. Provide \texttt{REASON}.\\
   - If valid: Output \texttt{STATUS: PASS}.
\end{tcolorbox}
\caption{The system prompt used in the Verification Stage of RoSci-Agent.}
 \Description{The prompt asks a scientific reviewer to check whether a problem can be solved. The reviewer extracts values, units and conditions from the text and data from the image, then checks for contradictions, violations of scientific laws and missing variables. A critical error or missing requirement calls for REJECT with a reason. A valid problem calls for PASS.}
 \label{fig:sysprompt}
 \vspace{-0.5cm}
\end{figure}
\begin{table}[t]
\centering
\begin{tabular}{lclc}
\toprule
\textbf{Category} & \textbf{Metric} & \textbf{Severity Level} & \textbf{Weight ($w_k$)} \\
\midrule
\multirow{4}{*}{Info Error} 
& MFE & Low (Minor Factual) & 0.3 \\
& CFE & Medium (Critical Error) & 0.7 \\
& VOL & High (Law Violation) & 1.0 \\
& CMC & High (Contradiction) & 1.0 \\
\midrule
\multirow{2}{*}{Info Loss} & CFL & High (Critical Factual Loss) & 1.0 \\
& CIL & High (Critical Image Loss) & 1.0 \\
\bottomrule
\end{tabular}
\caption{Weight parameters for RoSci-Agent.}
\vspace{-0.5cm}
\label{tab:weights}
\end{table}

\subsection{Weight Assignment for Anomaly Scoring}
\label{app:weights}
To ensure reproducibility, Table~\ref{tab:weights} details the severity weights used in Eq. \ref{eq:anamaly}. Fundamental violations that render a scientific problem unanswerable (e.g., VOL, CFL) are assigned higher weights.

\subsection{Findings of RoSci-Bench \& RoSci-Agent}
\begin{figure}[!htbp]
    \centering
    \begin{tcolorbox}[colback=purple!5, colframe=purple!75, title=\textbf{Baseline Model (Standard Inference)}]
    
    \textbf{Input:} Question: A car travels distance $d$. What is the speed? (Image shows a car). \\
    \textbf{Model Output:} 
    Based on the standard formula $v = d/t$. Assuming the time taken is 1 hour for typical highway travel, and estimating $d=60$ miles from context... \\
    \textbf{Final Answer:} The speed is 60 mph. \\
    \textbf{\textcolor{purple}{Analysis: Hallucination. The model invents specific values ($t=1h$) to force an answer.}}
    \end{tcolorbox}

    \begin{tcolorbox}[colback=blue!5, colframe=blue!60, title=\textbf{RoSci-Agent (Our Approach)}]
    
    \textbf{Phase 1: Knowledge Parsing} \\
    - Known: Distance $d$ (Variable). \\
    - Missing: Time $t$. \\
    \textbf{Phase 2: Verification Step} \\
    - Check: Can $v$ be calculated with only $d$? $\rightarrow$ NO. \\
    - Check: Is $t$ hidden in the image? $\rightarrow$ NO. \\
    \textbf{Phase 3: Decision} \\
    - Action: \texttt{REJECT}. \\
    \textbf{Model Output:} 
    The problem provides distance $d$ but fails to specify the time interval $t$. Speed is defined as $v=d/t$. Without $t$, the value cannot be determined. \\
    \textbf{\textcolor{blue!50!black}{Analysis: Correct Rejection. The agent identifies the missing condition before reasoning.}}
    \end{tcolorbox}
    \caption{Qualitative Comparison of Baseline model (top) and RoSci-Agent (bottom).}
    \Description{A car's travel distance is given symbolically. The baseline assumes one hour and estimates 60 miles, then reports 60 miles per hour. RoSci-Agent identifies time as missing, checks that the image does not supply it, and rejects the question because distance alone cannot determine speed.}
 \label{fig:qualitative_case}
\end{figure}

\begin{table}[!htbp]
    \centering
    \begin{tabular}{lccc}
        \toprule
        \textbf{Method} & {Avg. Tokens} & {Overall Score} & {Rejection F1} \\
        \midrule
        Standard (Zero-shot) & $\sim$520 & 30.21 & 47.16 \\
        Self-Consistency ($k=5$) & $\sim$2600 & 33.45 & 48.80 \\
        \rowcolor{myblue} \textbf{RoSci-Agent (Our)} & $\sim$1850 & \textbf{44.77} & \textbf{51.17} \\
        \bottomrule
    \end{tabular}
    \caption{Efficiency Comparison of RoSci-Agent and Self-Consistency baseline.}
    \label{tab:compute_cost}
\end{table}
\subsection{Additional Analyses}
\textbf{Efficiency Comparison.}
Table~\ref{tab:compute_cost} reports average token use, Overall Score, and Rejection F1 for direct answering, \textbf{Self-Consistency (SC)} with $k=5$ paths, and RoSci-Agent.

\textbf{Qualitative Analysis.}
To intuitively demonstrate the mechanism of RoSci-Agent, we visualize a comparison between a Standard VLM (direct inference) and our approach (CoT reasoning) on an \textit{Information Loss} case. As shown in Figure \ref{fig:qualitative_case}, the baseline assumes a travel time of one hour and a distance of 60 miles. In this example, RoSci-Agent checks the image for the missing time interval before rejecting the question.

\begin{table}[t]
    \centering
    \begin{tabular}{l|ccc|c}
        \toprule
        \textbf{Domain} & \textbf{Baseline} & \textbf{RoSci-Agent} & \textbf{Improvement} & \textbf{Sample Size} \\
        \midrule
        Physics & 28.5 & {46.2} & +17.7 & 104 \\
        Chemistry & 31.2 & {45.8} & +14.6 & 85 \\
        Biology & 35.0 & {43.5} & +8.5 & 76 \\
        Math & 25.4 & {41.0} & +15.6 & 40 \\
        \bottomrule
    \end{tabular}
    \caption{Robustness Score by Domain of RoSci-Agent.}
    \label{tab:domain_scores}
\end{table}

\textbf{Performance Breakdown across disciplines.}
\label{subsec:domain_breakdown}
Table~\ref{tab:domain_scores} shows the largest observed gain in physics (+17.7 points) and the smallest in biology (+8.5 points).

\end{document}